\documentclass{article}
\usepackage{spconf,amsmath,graphicx}

\usepackage{enumitem}
\setlist{nosep, leftmargin=14pt}

\usepackage{mwe} 
\usepackage{ dsfont }
\usepackage[percent]{overpic}
\usepackage{xcolor}

\title{Spherical Harmonics for Shape-Constrained 3D Cell Segmentation}
\name{Dennis Eschweiler$^{1}$, Malte Rethwisch$^{1}$, Simon Koppers$^{1}$, Johannes Stegmaier$^{1}$}
\address{$^{1}$ Institute of Imaging and Computer Vision, RWTH Aachen University, Aachen, Germany}
\begin{document}
%
\maketitle
\begin{abstract}
Recent microscopy imaging techniques allow to precisely analyze cell morphology in 3D image data.
To process the vast amount of image data generated by current digitized imaging techniques, automated approaches are demanded more than ever.
Segmentation approaches used for morphological analyses, however, are often prone to produce unnaturally shaped predictions, which in conclusion could lead to inaccurate experimental outcomes.
In order to minimize further manual interaction, shape priors help to constrain the predictions to the set of natural variations.
In this paper, we show how spherical harmonics can be used as an alternative way to inherently constrain the predictions of neural networks for the segmentation of cells in 3D microscopy image data.
Benefits and limitations of the spherical harmonic representation are analyzed and final results are compared to other state-of-the-art approaches on two different data sets.
\end{abstract}
\section{Introduction}
\label{sec:intro}
The continuous development of microscopy imaging techniques allows to better understand developmental processes at the cellular level.
Particularly 3D imaging techniques provide powerful insights, but create vast amounts of data that have to be analyzed, which renders manual investigations a tedious or even infeasible task.
Consequently, automated cell segmentation has been addressed with different machine learning-based approaches.
To leverage the efficacy of these automated approaches for biomedical assessments, the predicted segmentations need to be as accurate as possible.

Current instance segmentation approaches most commonly work in a pixel-wise manner and, \textit{e.g.}, utilize deep learning to predict multi-class segmentations, which are further processed to obtain individual instances \cite{guerrero2020, wolny2020, eschweiler2019a}, utilize deep learning to refine foreground proposals \cite{wolf2020, stegmaier2018} or use deep learning to directly predict positional feature maps \cite{stringer2020}.
Those techniques are reported to accurately distinguish separate cell instances, but, due to the pixel-wise working principle, generated segmentations are still prone to having fragmented or unnatural shapes.
To limit, yet not fully diminish, this problem of possible noisy segmentations, a huge amount of diverse training data needs to be used to allow the machine learning approaches to learn all possible variations.
The dilemma lies in the absence of those annotated data sets, which are seldom or not available at all, especially for large 3D image data.
To still constrain segmentations to natural shape variations, shapes can be represented by their global shape representation instead of a pixel-level representation.
Possible shape representations are Fibonacci lattices, which proofed to be applicable to microscopy image segmentation tasks \cite{weigert2019}, or spherical harmonics \cite{muller2006}, which have been shown to be suitable for a meaningful representation of 3D cell shapes \cite{ducroz2011, ducroz2012}.
Those shape representations have a trade-off between accuracy and complexity, which is one of the key factors that needs to be considered when deploying those methods.

The proposed work demonstrates, (1) how spherical harmonics can be used as an alternative way to predict inherently shape-constrained instance segmentations and (2) how this can be learned by current deep learning approaches.
We demonstrate, (3) that only a small amount of descriptors are required to accurately represent shapes and compare results to other methods established in the field.

\section{Predicting Spherical Harmonics}
\label{sec:motivation}
Using spherical harmonics (SH), every spherical shape $\mathbf{S}$ can be decomposed into $R$ different basis functions $Y_j$ weighted by scalars $c_j$.
An individual spherical shape is defined by 
\begin{equation}
   \small \mathbf{S} = \sum_{j=1}^{R} c_j \cdot Y_j\mathrm{,}
\end{equation}
where
\begin{equation}
    \small Y_j = Y_l^m(\theta,\phi) = \sqrt{\frac{2l + 1}{4 \pi} \cdot \frac{(l-m)!}{(l+m)!}} \cdot P_l^m(\cos \theta) e^{i\cdot m \cdot \phi}
\end{equation}
represents one basis function.
Here, $P_l^m(\cos \theta)$ describes the Legendre polynomials of degree $m$ and order $l$, with $l\geq0$ and $-l \leq m \leq l$, while higher orders encode the higher frequency components of the sphere.
Consequently, the first coefficient, \textit{i.e.}, the first basis function, represents a perfect sphere.
For each order $l$ we use all available degrees $m$, which allows to calculate the total number of basis functions up to order $l$ by
\begin{equation}
    R = (l+1)^2\mathrm{.} 
\end{equation}
To represent a spherical shape by only a small amount of SH basis functions, we need to limit the order $l$ and describe the shape by determining the weights $c_j$ for each corresponding basis function $Y_j$, with $j\in(0,R)$.
Parameters $\theta$ and $\phi$ denote the spherical angular sampling coordinates and their quantity further determines how detailed each shape can be represented.
Instead of sampling those angular orientations from a fixed uniform grid, we select them following the concept of electrostatic repulsion \cite{jones1999}.
This ensures an optimal sampling of each shape given a fixed number of orientations, which is empirically set to 5000 orientations to get an oversampled representation of each shape.

For the transformation between volumetric pixel-wise segmentations and spherical harmonics representations, each instance volume is transformed into a set of spherical vectors.
Originating at the centroid of each volume, radii are sampled along the angular orientations ($\theta, \phi$), resulting in a total of 5000 vectors $\mathcal{V}=(r, \theta, \phi)$.
Utilizing the predefined basis functions and the angular sampling pattern, the weights $c_j$ for each basis function can be determined, ultimately resulting in the spherical harmonic representation of each shape. 
To reverse the encoding to a pixel-wise representation, again the same angular sampling pattern is used to obtain the vectorized representation $\mathcal{V}$.
Subsequently, each volume is reconstructed by applying a Delaunay triangulation.

\begin{figure}[h]
    \centering
    \includegraphics[width=0.48\textwidth]{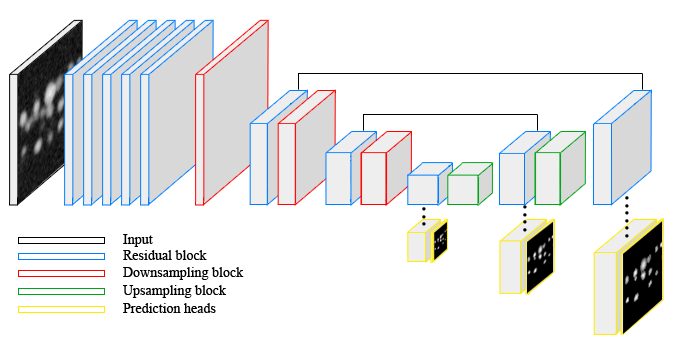}
    \caption{Illustration of the shape prediction network, which comprises a residual-based encoder-decoder architecture to make predictions at three different scales. Skip connections combine low-resolution and high-resolution features. Each prediction head provides spherical harmonic encodings at each pixel position and a distance map.} 
    \label{fig:network}
\end{figure}

The proposed HarmonicNet architecture is inspired by the work done in \cite{redmon2018} and \cite{weigert2019} and comprises a residual-based encoder and decoder part, making predictions at three different scales and using skip-connections to combine low-resolution and high-resolution features (Fig. \ref{fig:network}).
Each block consists of three $3\times3\times3$ convolutional layers with one additional $1\times1\times1$ convolutional layer for each prediction head.
Predictions at each scale are given by downsampling the input dimensions by factor 8, 4 and 2 respectively.
Note that the prediction with the highest resolution downsamples the input dimension by factor two to reduce the output size by still maintaining a high level of detail.
The prediction heads are split into two paths for prediction of SH coefficients at each pixel position and a distance map.
The distance map shows the relative distance to the boundary for each pixel within an instance and zero for background pixel.
Spherical harmonic encodings for each shape are given at each pixel position within the respective instance.
The number of output channels depends on the chosen total number of SH basis functions $R$, resulting in a total of $1+R$ predictions per pixel at each scale.
Since SH coefficient values are theoretically unbounded in $\mathds{R}$ while the distance map values are in $(0,1)$, PReLU activation functions \cite{he2015} are used after each convolutional layer, allowing the network to dynamically learn the negative slope.
Furthermore, the final output activation for each spherical harmonic coefficient prediction path is omitted to allow unbounded outputs and a sigmoid activation function is used at the end of each detection path.

Two different L1 loss terms are combined to formulate the training loss.
The first loss assesses the predicted distance maps by weighting the L1 scores at each pixel position by the occurrence of foreground and background pixels, which can be formulated as
\begin{equation}
    \label{equ:loss_distance}
    \mathcal{L}_{dist}(\mathbf{x}_{t},\mathbf{x}_{p}) = \left(\frac{\mathbf{x}_{t\geq 0.5}}{\sum \mathbf{x}_{t\geq 0.5}} + \frac{\mathbf{x}_{t< 0.5}}{\sum \mathbf{x}_{t< 0.5}}\right)\cdot ||\mathbf{x}_{t}-\mathbf{x}_{p}||\mathrm{,}
\end{equation}
with $\mathbf{x}_{t}$ and $\mathbf{x}_{p}$ representing the ground truth and prediction maps, respectively.
The second loss assesses the predicted SH encodings by considering only predictions within the foreground region, as there are no encodings defined within the background.
This can be formulated as
\begin{equation}
    \label{equ:loss_encoding}
    \mathcal{L}_{harm}(\mathbf{y}_{t},\mathbf{y}_{p},\mathbf{x}_{t}) = \mathbf{x}_{t>0}\cdot ||\mathbf{y}_{t}-\mathbf{y}_{p}||\mathrm{,}
\end{equation}
with $\mathbf{y}_{t}$ and $\mathbf{y}_{p}$ representing the ground truth and prediction encodings, respectively.
The final loss is given as combination of both terms, weighted by scalars $\lambda_{dist}$ and $\lambda_{harm}$:
\begin{equation}
    \label{equ:loss_combined}
    \mathcal{L} = \lambda_{dist}\cdot\mathcal{L}_{dist} + \lambda_{harm}\cdot\mathcal{L}_{harm}\mathrm{.}
\end{equation}

To obtain the instance segmentation, centroid locations are detected by finding peaks in the predicted distance maps, while peaks are defined as the local maxima within a region of $2\cdot d_{min}$ exceeding an intensity threshold of $t_{det}$. 
Each distance map was scaled up to the input size before peaks were detected, which helps to adjust the region parameter $d_{min}$ to the observed cell sizes.
Within a small $5\times5\times5$ neighbourhood around each detected centroid, spherical harmonic encodings are weighted by their predicted relative distance to the centroid and averaged to obtain the final robust shape encoding.
These encodings are transformed to the vectorized representation $\mathcal{V}$ and, subsequently, a Delaunay triangulation is applied to retrieve the pixel-wise instance volume.
\begin{figure*}[h]
    \centering
    \begin{overpic}[height=0.232\textwidth]{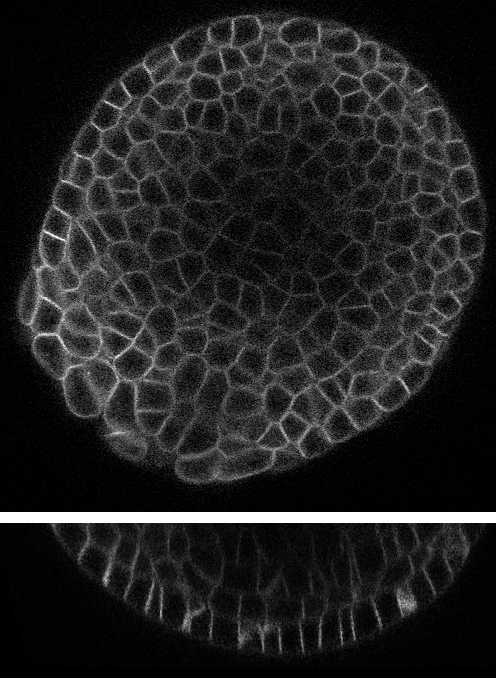} \put(3,92){\scriptsize\color{white} XY} \put(3,3){\scriptsize\color{white} XZ} \end{overpic}
    \begin{overpic}[height=0.232\textwidth]{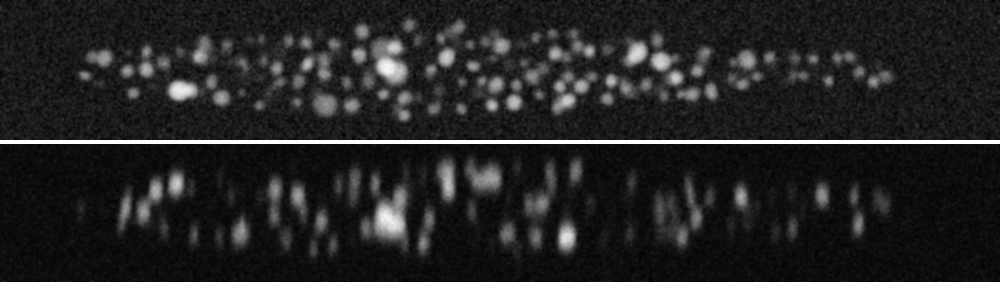} \put(1,26){\scriptsize\color{white} XY} \put(1,1){\scriptsize\color{white} XZ} \end{overpic}
    \caption{Sample images from the meristem data set (left) \cite{willis2016} and the simulated cell nuclei (right). The upper row shows slices from the XY plane and the lower row shows slices from the XZ plane.} 
    \label{fig:images}
\end{figure*}

\section{Experiments and Results}
\label{sec:exp}
For the following experiments, two different data sets were used.
The first data set comprised 125 3D confocal microscopy images showing fluorescently labeled cell membranes of \textit{Arabidopsis thaliana} and corresponding manually corrected watershed-based instance segmentations \cite{willis2016}.
Plant 2, 4, 8 and 13 were used for training and plant 15 and 18 were used for testing.
The second data set comprised 60 artificially generated 3D nuclei images.
Spherical harmonics were used to place roughly 350 individual cells into images of size $1000\times140\times140$ pixel.
Subsequently, two measured point spread functions (PSF) \cite{preibisch2014} and additive Gaussian noise $\mathcal{N}(0,0.1)$ were used to generate realistic image distortions.
The training set was composed of 20 images distorted by additive noise only and 20 images distorted by the first PSF plus additive noise.
For training the remaining 20 images were simulated by the second PSF plus additive noise to obtain structural differences between the training and test set.

\begin{figure}[b!]
    \centering
    \includegraphics[width=0.49\textwidth]{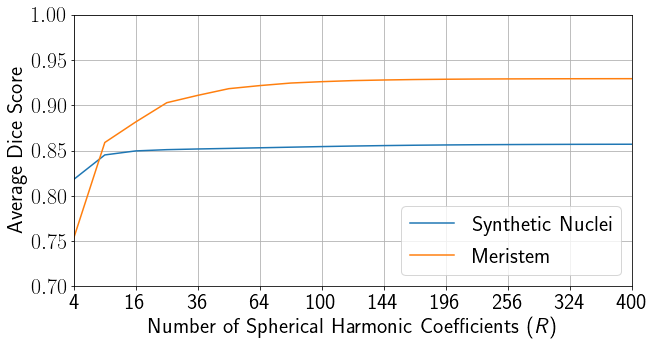}
    \caption{Average Dice scores obtained after encoding and reconstructing segmentation instances from both data sets. We chose 36 coefficients as a good trade-off between accuracy and complexity.} 
    \label{fig:tradeoff}
\end{figure}

To appraise the segmentation results obtained with the proposed method, we first evaluated the trade-off between accuracy and complexity of the spherical harmonic encoding.
For shape prediction, the spherical harmonic order $l$ and, thereby, the number of spherical harmonic coefficients $R$ has to be chosen, which also defines how precise high-frequency information can be represented.
The maximum possible segmentation scores that can be obtained on both data sets using different $R$ is shown in Figure \ref{fig:tradeoff}.

Results obtained by the proposed network were compared to three different methods used in the field, namely 3D StarDist \cite{weigert2019}, a 3D variant of Cellpose \cite{stringer2020} and a 3D UNet-assisted watershed-based segmentation (UNet+WS) \cite{eschweiler2019a}.
For better comparability, we set the number of rays used by the StarDist approach to 36.
The Cellpose approach was adapted to work in a full 3D manner, as we predicted gradient fields in x,y and z direction and the additional foreground map.
Furthermore, the definition of the gradient fields was simplified by spanning a tanh function between both sides of each shape in each respective direction.
During the iterative reconstruction, each position was moved one voxel per iteration along the predicted 3D gradient and the total number of iterations was set to the estimated maximum size of cells.
Cell instances were obtained by a final clustering of positions.

\begin{figure}[b!]
    \centering
    \includegraphics[width=0.23\textwidth]{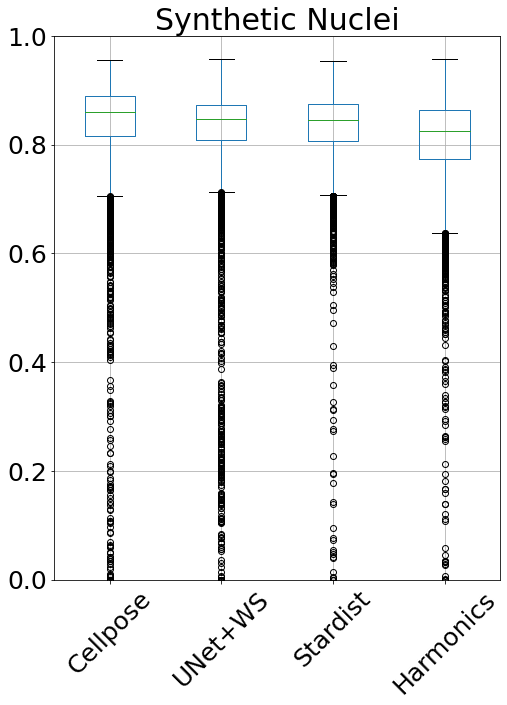}
    \includegraphics[width=0.23\textwidth]{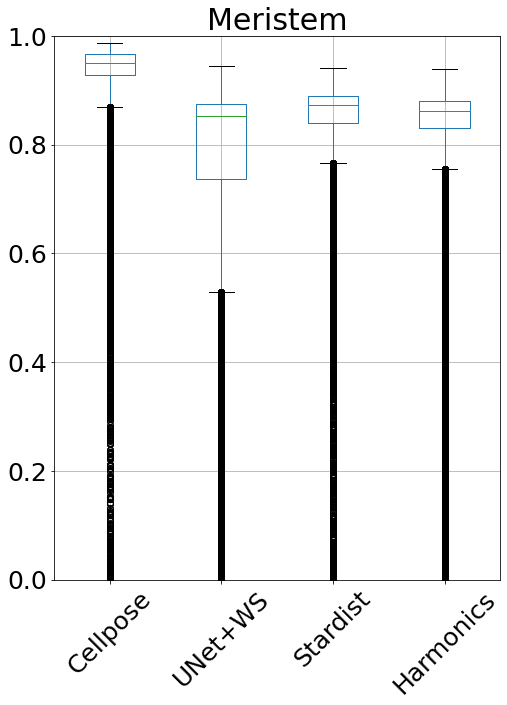}
    \caption{Boxplots showing averaged Dice scores obtained for each method on both the nuclei and the meristem data set. Outliers are marked as circles.} 
    \label{fig:results}
\end{figure}
\begin{figure*}[ht!]
    \centering
    \begin{overpic}[height=0.1465\textwidth]{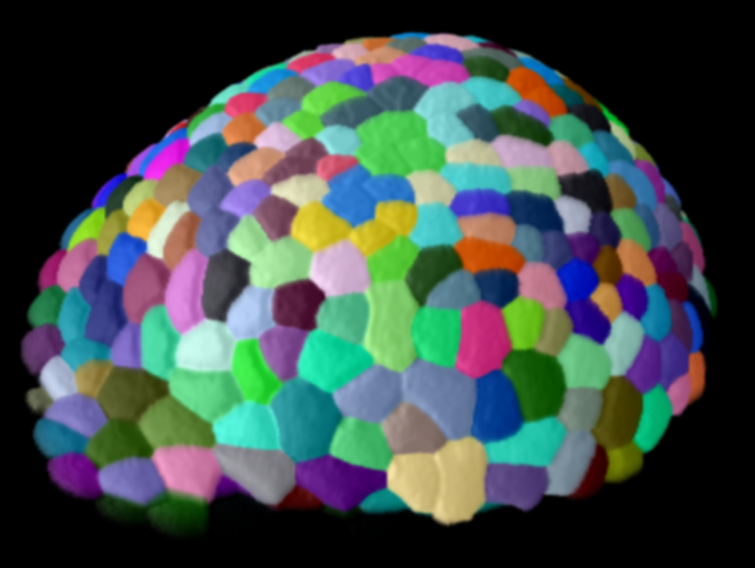} \end{overpic}
    \begin{overpic}[height=0.1465\textwidth]{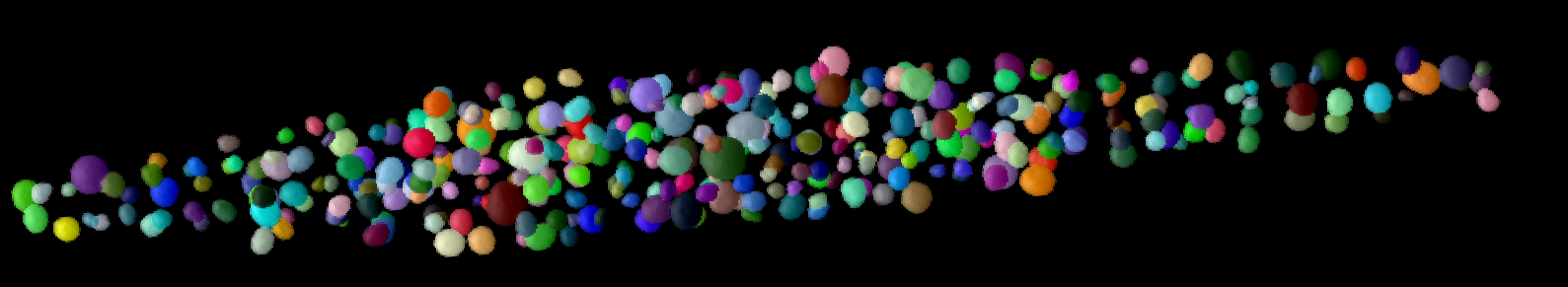} \put(1,16){\scriptsize\color{white} Ground Truth} \end{overpic}
    \begin{overpic}[height=0.1465\textwidth]{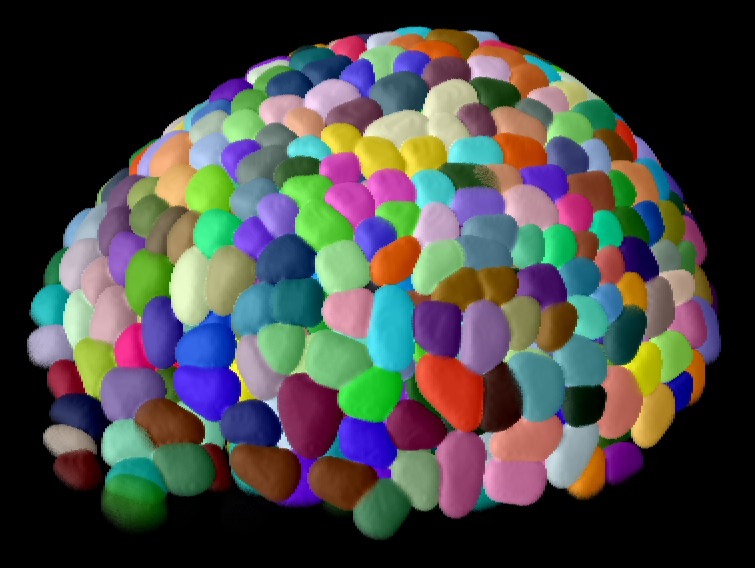} \end{overpic}
    \begin{overpic}[height=0.1465\textwidth]{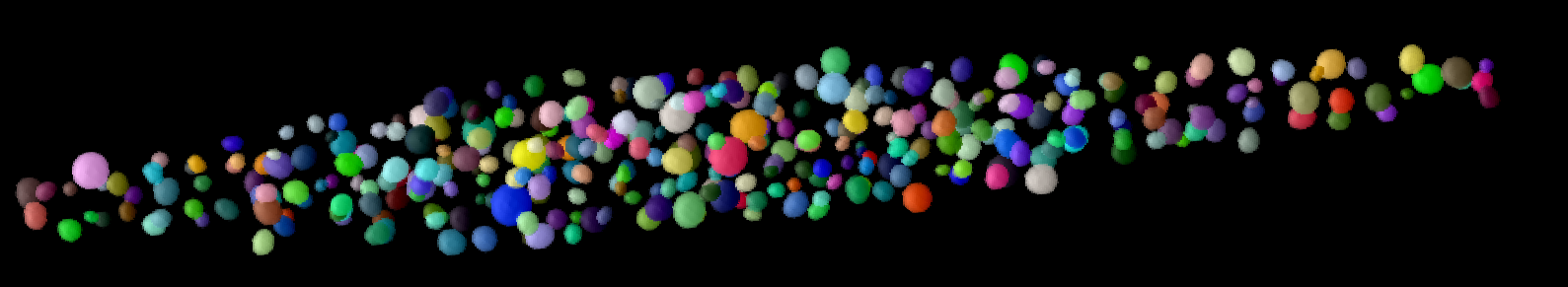} \put(1,16){\scriptsize\color{white} Prediction} \end{overpic}
    \caption{3D renderings of the instance segmentation obtained by the spherical harmonics approach for the meristem data set (left) and for the simulated nuclei (right). The first row shows the ground truth and the second row shows the predictions.} 
    \label{fig:predictions}
\end{figure*}
Considering the results shown in Figure \ref{fig:tradeoff}, we chose SH order $l=5$, \textit{i.e.} $R=36$, as a good balance between accuracy and number of output parameters.
This results in potentially reachable averaged Dice scores of 0.91 for the meristem data set and 0.85 for the simulated nuclei data set.
During training, both loss terms contributed equally by setting $\lambda_{dist}$ and $\lambda_{harm}$ to $0.5$.
For the determination of peaks, the probability threshold $t_{det}$ was set to $0.5$ and the region size $d_{min}$ was set to $20$ for the meristem data set and $10$ for the simulated nuclei data set.
For evaluation of the instance segmentation we employed an averaged instance-level Dice score, which is shown in Figure \ref{fig:results}.

Experiments on the nuclei data set assess how accurate small cells can be segmented.
Results show that all approaches perform similarly well and errors are mainly caused by small offsets between ground truth shapes and predictions.
Note that both shape-level approaches, \textit{i.e.}, StarDist and the spherical harmonics approach, are reported to have a maximum reachable accuracy due to the loss of high-frequency shape components. 
Despite this disadvantage, they are on par with both pixel-level approaches, \textit{i.e.}, Cellpose and UNet+WS.
Particularly the proposed pipeline achieves a median averaged Dice score of 0.83, which almost reaches the maximum obtainable score of 0.85 for 36 spherical harmonic coefficients (Fig. \ref{fig:tradeoff}).
Median scores of the other approaches range between 0.84 and 0.86 (Fig. \ref{fig:results}).

\begin{figure}[h]
    \centering
    \begin{overpic}[width=0.093\textwidth]{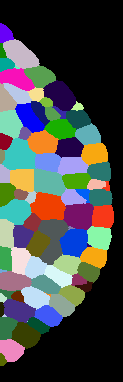} \put(1,95){\scriptsize\color{white} Ground Truth} \end{overpic}
    \begin{overpic}[width=0.093\textwidth]{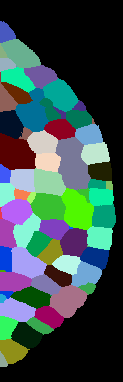} \put(1,95){\scriptsize\color{white} Cellpose} \end{overpic}
    \begin{overpic}[width=0.093\textwidth]{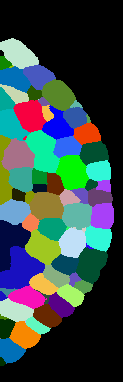} \put(1,95){\scriptsize\color{white} UNet+WS} \end{overpic}
    \begin{overpic}[width=0.093\textwidth]{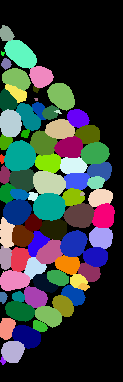} \put(1,95){\scriptsize\color{white} StarDist} \end{overpic}
    \begin{overpic}[width=0.093\textwidth]{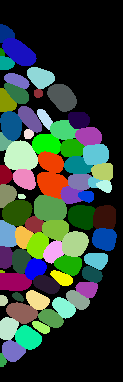} \put(1,95){\scriptsize\color{white} HarmonicNet} \end{overpic}
    \caption{2D crops of the ground truth and segmentations obtained by Cellpose, UNet+WS, StarDist and HarmonicNet.} 
    \label{fig:segerror}
\end{figure}

Experiments on the meristem data set assess how well each of the examined approaches performs when applied to very dense cell populations.
Results show a discrepancy between both pixel-level approaches with Cellpose achieving a median averaged Dice of 0.95 and the UNet+WS approach achieving a median score of 0.85.
This exposes the fragility of pixel-level approaches and the necessity of accurate network outputs to obtain accurate and natural shapes.
The shape-level approaches reach similar results as for the nuclei data set, ranging between median averaged dice scores of 0.87 for StarDist and 0.87 for the SH approach, which demonstrates the characteristics of the shape encoding, as they limit shapes to their natural variations and thereby prevent larger errors.
A comparison of segmentations obtained by each approach is visualized in Figure \ref{fig:segerror}.

\section{Conclusion}
\label{sec:motivation}
In this paper we demonstrated that spherical harmonics can be utilized to predict shape-constrained segmentations with current deep learning approaches.
The proposed pipeline proved to be competitive to other approaches used in the field, rendering it a good alternative to existing methods.
As spherical harmonics are a natural way to represent spherical cell shapes, they offer high shape flexibility by using only a small amount of coefficients.
Depending on the complexity of the data set, the number of descriptors could be further reduced without a significant loss of reachable accuracy, resulting in a robust and parameter-efficient way to obtain segmentations. 
Additionally, they could be combined with pixel-level approaches to incorporate further high-frequency shape information.

%
%
%
%
%
%


\bibliographystyle{IEEEbib}
\bibliography{refs}

\begin{thebibliography}{10}

\bibitem{guerrero2020}
F.~A. Guerrero-Pe{\~n}a, P.~D.~M. Fernandez, P.~T. Tarr, et~al.,
\newblock ``{J Regularization Improves Imbalanced Multiclass Segmentation}'',
\newblock in {\em IEEE International Symposium on Biomedical Imaging (ISBI)},
  2020, pp. 1--5.

\bibitem{wolny2020}
A.~Wolny, L.~Cerrone, A.~Vijayan, et~al.,
\newblock ``{Accurate and Versatile 3D Segmentation of Plant Tissues at
  Cellular Resolution}'',
\newblock {\em bioRxiv}, 2020.

\bibitem{eschweiler2019a}
D.~Eschweiler, T.~V. Spina, R.~C. Choudhury, et~al.,
\newblock ``{CNN-based Preprocessing to Optimize Watershed-based Cell
  Segmentation in 3D Confocal Microscopy Images}'',
\newblock in {\em International Symposium on Biomedical Imaging (ISBI)}, 2019,
  pp. 223--227.

\bibitem{wolf2020}
S.~Wolf, F.~A. Hamprecht, and J.~Funke,
\newblock ``{Inpainting Networks Learn to Separate Cells in Microscopy
  Images}'',
\newblock in {\em The British Machine Vision Conference (BMVC)}, 2020.

\bibitem{stegmaier2018}
J.~Stegmaier, T.~V. Spina, A.~X. Falc{\~a}o, et~al.,
\newblock ``{Cell Segmentation in 3D Confocal Images Using Supervoxel
  Merge-Forests with CNN-based Hypothesis Selection}'',
\newblock in {\em International Symposium on Biomedical Imaging (ISBI)}, 2018,
  pp. 382--386.

\bibitem{stringer2020}
C.~Stringer, M.~Michaelos, and M.~Pachitariu,
\newblock ``{Cellpose: a Generalist Algorithm for Cellular Segmentation}'',
\newblock {\em bioRxiv}, 2020.

\bibitem{weigert2019}
M.~Weigert, U.~Schmidt, R.~Haase, K.~Sugawara, and G.~Myers,
\newblock ``{Star-Convex Polyhedra for 3D Object Detection and Segmentation in
  Microscopy}'',
\newblock in {\em IEEE Winter Conference on Applications of Computer Vision
  (WACV)}, 2020, pp. 3666--3673.

\bibitem{muller2006}
C.~M{\"u}ller,
\newblock {\em {Spherical Harmonics}}, vol.~17,
\newblock Springer, 2006.

\bibitem{ducroz2011}
C.~Ducroz, J.-C. Olivo-Marin, and A.~Dufour,
\newblock ``{Spherical Harmonics based Extraction and Annotation of Cell Shape
  in 3D Time-Lapse Microscopy Sequences}'',
\newblock in {\em International Conference of the IEEE Engineering in Medicine
  and Biology Society}, 2011, pp. 6619--6622.

\bibitem{ducroz2012}
C.~Ducroz, J.-C. Olivo-Marin, and A.~Dufour,
\newblock ``{Characterization of Cell Shape and Deformation in 3D using
  Spherical Harmonics}'',
\newblock in {\em IEEE International Symposium on Biomedical Imaging (ISBI)},
  2012, pp. 848--851.

\bibitem{jones1999}
D.~K. Jones, M.~A. Horsfield, and A.~Simmons,
\newblock ``{Optimal Strategies for Measuring Diffusion in Anisotropic Systems
  by Magnetic Resonance Imaging}'',
\newblock {\em Magnetic Resonance in Medicine: An Official Journal of the
  International Society for Magnetic Resonance in Medicine}, vol. 42, pp.
  515--525, 1999.

\bibitem{redmon2018}
J.~Redmon and A.~Farhadi,
\newblock ``{Yolov3: An Incremental Improvement}'',
\newblock {\em arXiv:1804.02767}, 2018.

\bibitem{he2015}
K.~He, X.~Zhang, S.~Ren, and J.~Sun,
\newblock ``{Delving Deep Into Rectifiers: Surpassing Human-Level Performance
  on ImageNet Classification}'',
\newblock in {\em IEEE International Conference on Computer Vision (ICCV)},
  2015, pp. 1026--1034.

\bibitem{willis2016}
L.~Willis, Y.~Refahi, R.~Wightman, et~al.,
\newblock ``{Cell Size and Growth Regulation in the Arabidopsis Thaliana Apical
  Stem Cell Niche}'',
\newblock {\em Proceedings of the National Academy of Sciences}, vol. 113, no.
  51, pp. E8238--E8246, 2016.

\bibitem{preibisch2014}
S.~Preibisch, F.~Amat, E.~Stamataki, et~al.,
\newblock ``{Efficient Bayesian Multi-View Deconvolution}'',
\newblock in {\em Nature Methods}, 2014, pp. 645--648.

\end{thebibliography}

\end{document}